\pdfoutput=1

\documentclass[11pt]{article}

\usepackage{acl2023}
\usepackage{times}
\usepackage{latexsym}
\usepackage{multirow}
\usepackage[T1]{fontenc}

\usepackage[utf8]{inputenc}

\usepackage{microtype}

\usepackage{graphicx}
\usepackage{multirow}
\usepackage{amsmath}
\usepackage{amssymb}

\usepackage{color, colortbl}
\usepackage{soul}

\title{Enhancing Distractor Generation for Multiple-Choice Questions with Retrieval Augmented Pretraining and Knowledge Graph Integration}

\author{Han-Cheng Yu, Yu-An Shih, Kin-Man Law, Kai-Yu Hsieh, \\ \textbf{Yu-Chen Cheng}, \textbf{Hsin-Chih Ho}, \textbf{Zih-An Lin}, \textbf{Wen-Chuan Hsu}, \\ \textbf{Yao-Chung Fan$^*$}\\
  Department of Computer Science and Engineering, \\
  National Chung Hsing University, Taiwan\\
  \texttt{yfan@nchu.edu.tw}
  }

\begin{document}
\maketitle
\begin{abstract}
In this paper, we tackle the task of distractor generation (DG) for multiple-choice questions. Our study introduces two key designs. First, we propose \textit{retrieval augmented pretraining}, which involves refining the language model pretraining to align it more closely with the downstream task of DG. Second, we explore the integration of knowledge graphs to enhance the performance of DG. Through experiments with benchmarking datasets, we show that our models significantly outperform the state-of-the-art results. Our best-performing model advances the F1@3 score from 14.80 to 16.47 in MCQ dataset and from 15.92 to 16.50 in Sciq dataset.

\end{abstract}

\section{Introduction}


Multiple-Choice Questions (MCQs) are widely used to evaluate a learner's knowledge. However, creating them manually requires a significant amount of time and effort from educators. Well-designed MCQs help in accurately assessing learners' abilities. The most challenging part of creating MCQs is designing appropriate incorrect options, commonly called \textit{distractors}. Therefore, researchers have focused on the automatic distractor generation for MCQs in recent years. In this paper, we explore two directions for improving DG methods.

\noindent\textbf{Task-Specific Pretraining} Current SOTA DG methods, such as utilizing BERT for DG \cite{chiangCDGP} or harnessing T5 for Text2Text DG \cite{juan-2023-PKLD}, are all based on pre-trained language models (LMs). Existing LMs undergo task-agnostic pertaining. However, recently, the idea of \textit{task-specific pretraining} has gained prominence. The aim is to refine the pretraining process to closely align with the downstream task. Therefore, our first endeavor in this paper is to investigate Task-Specific Pretraining for the DG task. 

In line with this goal, we propose \textit{Retrieval Augmented Pretraining} (RAP) for task-specific pretraining. The main idea involves using MCQ answers to retrieve relevant sentences/passages from a large corpus, such as Wikipedia, to create \textit{pseudo} questions, and then use the generated pseudo question for task-specific pertaining. 
    
\begin{figure}[t]
    \centering
    \includegraphics[width=\linewidth]{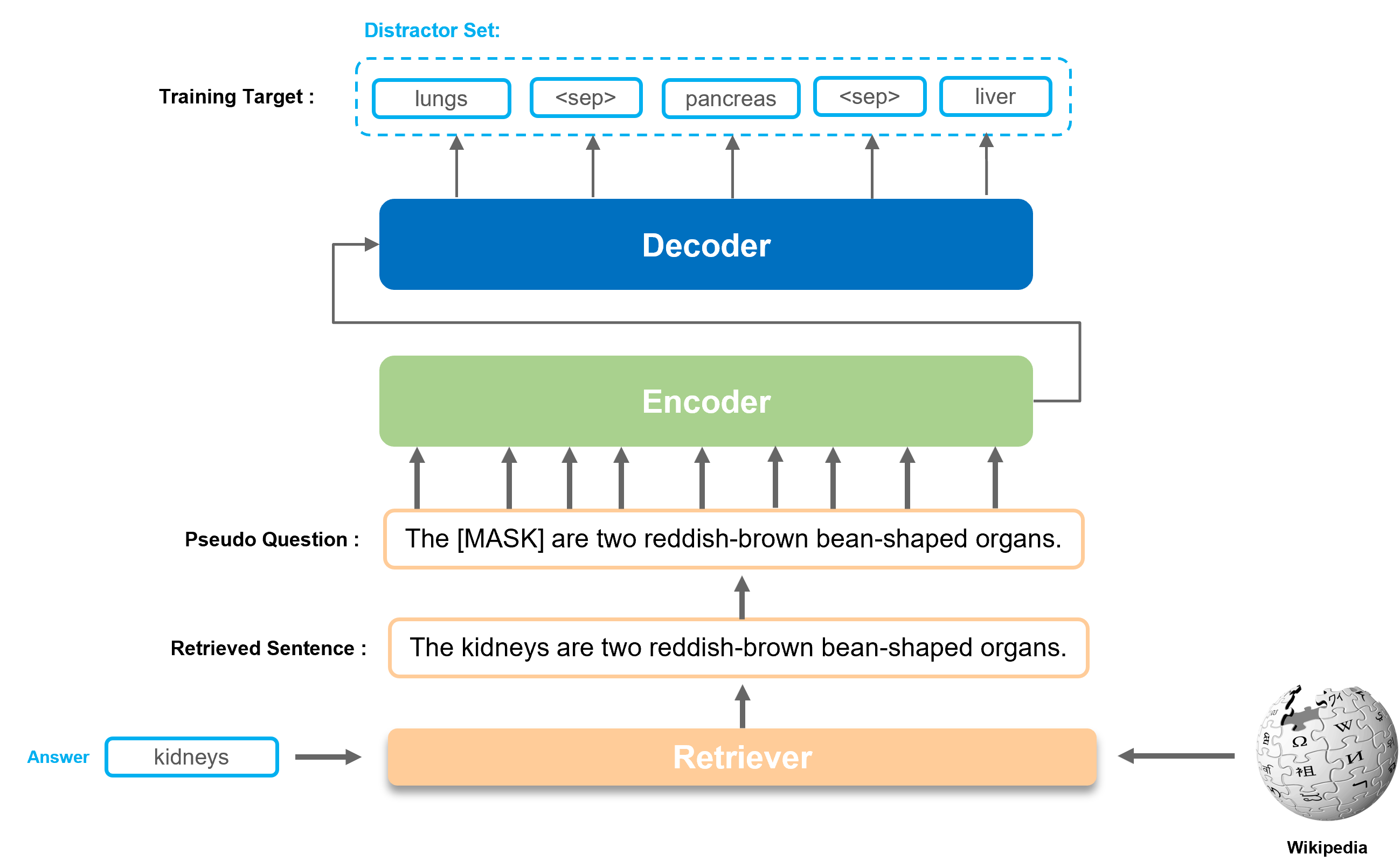}
    \caption{Retrieved Augmented Pretraining}
    \label{fig:SLGeneration}
\end{figure}

Figure \ref{fig:SLGeneration} illustrates an example of a create pseudo question. Assume that we are given a set of MCQ options, the answer \textit{kidneys}, and three distractors \textit{lungs, pancreas, liver}. The idea is to use the answer option \textit{kidneys} to retrieve a sentence, i.e., \textit{the kidneys are two reddish-brown bean-shaped organs.} and replace the answer with the $\mathtt{[Mask]}$ token to generate a pseudo question, i.e., \textit{the $\mathtt{[Mask]}$ are two reddish-brown bean-shaped organs}. The pseudo question and the distractors are then served as training data for RAP.

\noindent\textbf{Knowledge Augmented Generation} Recent research trends show that leveraging knowledge graphs (KGs) can enhance the performance of LM-based text generation tasks, as discussed in \cite{yasunaga2022deep} and \cite{zhang2022greaselm}. The perspective is that LMs and KGs should complement each other. KGs go beyond text by offering structural information, representing entities as nodes and their relationships as edges. This structured knowledge enhances the ability for multi-step reasoning. In line with this motivation, we propose \textit{knowledge-augmented generation} (KAG), where we build upon the candidate augmentation strategy proposed by the state-of-the-art DG method \cite{juan-2023-PKLD}. The idea is to retrieve knowledge triplets from a knowledge graph to serve as auxiliary information for the text2text DG model.

The contributions of this paper are 
\begin{itemize}

    \item Our methods achieve a remarkable improvement in state-of-the-art DG results. Our best-performing model elevates the F1@3 score from 14.80 to 16.47 in MCQ and from 15.92 to 16.50 in Sciq, showcasing the effectiveness of our approach.

    
    \item Extensive experimental evaluation with the benchmarking datasets are conducted and the insights of incorporating task-pretraining and knowledge triplet provision are discussed. Our study unveils promising directions for further development in DG by showcasing the efficacy of knowledge augmentation and task-specific pretraining.

    
\end{itemize}


The rest of this paper is organized as follows. Section \ref{sec:relatedWork} reviews the works of automatic distractor generation in the literature. In Section \ref{sec:methodology} we present the proposed methods. Section \ref{sec:experiment} reports the performance evaluation and Section \ref{sec:conclusion} concludes this work and discuss the future work.

\section{Related Work}
\label{sec:relatedWork} 
\subsection{Distractor Generation}
The current DG research can be divided into two directions:

\noindent\textbf{Generating and Ranking (GR) framework}
The framework incorporates a general-purpose knowledge base to effectively create a small distractor candidate set, and a feature-rich learning-to-rank model to select distractors. Specifically, the GR architecture consists of two stages.

First, it generates candidate distractors, and then it ranks these candidates based on semantic rules and linguistic features to select the final distractor. In the GR framework, there are two approaches for generating distractor candidates: using a knowledge base \cite{ren2021knowledge} or using a language model (LM) \cite{chiangCDGP}. These approaches have significantly improved the quality and diversity of distractors compared to traditional rule-based methods \cite{liang2017distractor,liang2018distractor}, making them the state-of-the-art in DG.

    \noindent\textbf{Text2Text generation architecture} 
    The Text2Text generation architecture differs from the GR architecture by formulating distractor generation as a Text2Text task. Specifically, it concatenates the question stem with the answer and inputs it into a generative language model (e.g., T5 or GPT) to train the model to generate a set of distractors. Currently, research based on the Text2Text architecture \citep{juan-2023-PKLD} represents the state-of-the-art in distractor generation.

\subsection{Task-specific Pretraining}
In recent related work, researchers have explored the integration of task-specific priors into BERT language model pretraining to enhance performance in low-resource finetuning tasks \cite{wang2020integrating}. Traditional pretraining methods often prioritize generic natural language knowledge, overlooking task-specific information, which can lead to overfitting in low-resource downstream tasks. To address this issue, \cite{wang2020integrating} propose integrating task-specific label embeddings into the self-attention layers during pretraining. This integration enables the model to filter out task-irrelevant information and improve task-specific knowledge during fine-tuning, resulting in reduced overfitting and improved performance on downstream tasks with limited resources.

Another related work, \cite{zhang2020pegasus}, adopts a task-specific pretraining approach for text summarization tasks. The authors replace a portion of the original sentences in the text with mask tokens, allowing the model to learn to generate masked sentences. Comparing this approach to traditional masked language modeling (MLM) for pretraining, they find that closer alignment between the pretraining task and the downstream task leads to improved performance. Additionally, they experiment with pretraining on various types of datasets and observe that utilizing a dataset similar to the downstream task yields better results. For example, when pretraining on a news dataset and fine-tuning on another news dataset, the performance surpasses that of fine-tuning on a web article dataset.

Building upon these findings, we draw inspiration from the aforementioned studies to design our proposed approach, known as RAP (Retrieval-Augmented Pretraining), aimed at enhancing the performance of distractor generation (DG). RAP leverages task-specific priors and incorporates them into the pretraining phase to capture and utilize relevant information specific to the distractor generation task. By aligning the pretraining task with the downstream DG task and considering task-specific information, we aim to improve the quality and effectiveness of distractor generation.

\subsection{Knowledge Augmented Generation}
We have conducted a survey of research that leverages Knowledge Graphs (KGs) to enhance performance in various Natural Language Processing (NLP) tasks, including Question Answering, Entity Typing/Relation Classification, Query Answering, Question Generation, and Distractor Generation \citep{yasunaga2022deep,yasunaga2021qa,zhang2022greaselm,feng2020scalable,he2019integrating,ren2020query2box,fei2022cqg,chen2023toward,ren2021knowledge}.

Notably, a significant amount of research has focused on using Knowledge Graphs to improve question-answering systems, particularly for tasks involving multihop reasoning. These studies utilize Knowledge Graph triplets for inferencing and have demonstrated the utility of structured background knowledge in strengthening NLP tasks.

For example, \citep{yasunaga2021qa} proposed an end-to-end question-answering model that combines language models and Knowledge Graphs to handle reasoning tasks. \citep{zhang2022greaselm} introduced a novel model that incorporates joint information exchange between language models and Knowledge Graphs, achieving superior performance across multiple domains.

While Knowledge Graph integration has shown promising results in Question-Answering tasks, its application to Distractor Generation remains relatively limited \citep{ren2021knowledge}.

Therefore, our goal is to leverage KGs as auxiliary tools to enhance distractor generation, improving the relevance of generated options to the question. This approach not only creates more challenging options for multiple-choice questions but also enhances the model's efficiency in utilizing Knowledge Graphs, leading to more effective Knowledge Augmented Generation.

\section{Methodology}\label{sec:methodology}
\subsection{Retrieval Augmented Pretraining}\label{sec:rap}
Our approach involves generating \textit{pseudo questions} to continuously train a pre-trained language model (e.g., T5 or BART) before fine-tuning it for downstream Distractor Generation (DG) tasks.

Here's how our idea works: Given an MCQ training instance consisting of a question stem ($q$), an answer option ($a$), and a set of distractors ($D$), we utilize $a$ to retrieve a single sentence or a short passage from a corpus (such as Wikipedia). Subsequently, we mask out $a$ from the retrieved sentence/passage, replacing the answer position with a mask token ($[\mathtt{Mask}]$). This process results in a modified sentence/passage known as a pseudo question ($\tilde{q}_{\otimes [\mathtt{Mask}]}$). 

With $\tilde{q}_{\otimes [\mathtt{Mask}]}$, our goal is to pretrain a LM as follows.
\[L(\theta) = -\sum_{i=1}^{|D|} t_{i} \log p(\hat{t}_{i} | \hat{t}_{<i}, \tilde{q}_{\otimes [\mathtt{Mask}]}, a; \theta)\]
\begin{itemize}
    \item $t_i$: the token sequence given by the concatenation of the ground truth distractors $d_1||d_2||d_3$.
\end{itemize}

\paragraph{Alternatives for RAP Training Setting}
There are two alternatives for RAP training setting. 
\begin{itemize}
    \item \textit{Data Augmentation RAP}: We can treat RAP as a data augmentation mechanism by using the same dataset for RAP and DG.
    \item \textit{Cross-Domain RAP}: We can use two different distractor generation datasets (SciQ and MCQ datasets in this study). One dataset is used for RAP and the other for DG fine-tuning. 
\end{itemize}

There are different impacts and implications for RAP training adoption. We provide experimental study and discussion for this part in Subsection \ref{sec:cross}.

\paragraph{Boosting RAP with ground-truth distractor}
The RAP idea can be extended to using ground-truth distractors to generate pseudo questions. That is, we can use ground-truth distractors to retrieve a sentence/passage and set the rest of options (i.e., $a$ and other distractors). That is,  
\[L(\theta) = -\sum_{i=1}^{|D|} t_{i} \log p(\hat{t}_{i} | \hat{t}_{<i}, \tilde{q}_{\otimes [\mathtt{Mask}]}, d_1; \theta)\]
\begin{itemize}
    \item $t_i$: the token sequence given by the concatenation of the other options $a||d_2||d_3$.
\end{itemize}

\subsection{Knowledge Augmented Generation}\label{sec:ca}
\begin{figure*}[t]
    \centering
    \includegraphics[width=\linewidth]{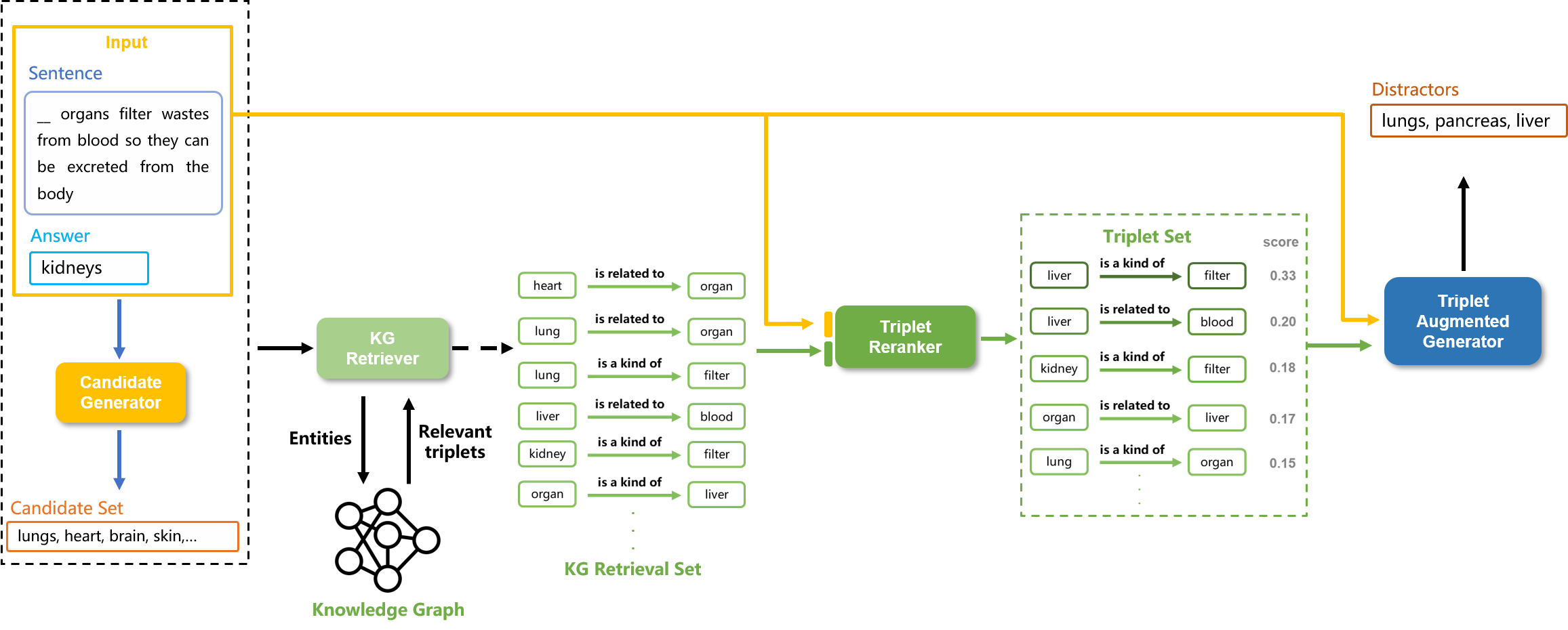}
    \caption{Knowledge Augmented Generation}
    \label{fig:Method-Overview}
\end{figure*}

We extend the idea of the candidate augmentation strategy proposed by \cite{juan-2023-PKLD}. We explore incorporating knowledge triplets for DG. Specifically, as shown in Figure \ref{fig:Method-Overview}, our KAG proceeds by (1) a knowledge triplet retrieval stage (Subsection \ref{sec:Retriever}) and (2) a triplet re-ranker stage (Subsection \ref{sec:ranker}). The retriever stage is to retrieve triplets with respect to $q$ and $a$, and the re-ranker stage aims to rank the triplets by their estimated relevancy with respect to $q$ and $a$

\subsubsection{Retrieving Triplet from KG}\label{sec:Retriever}
Our process for obtaining knowledge triplets is as follows. First, for a given $q$ and answer $a$, we employ an language model trained for distractor generation (e.g., \cite{chiangCDGP}) to generate a set of candidate distractors $\{\hat{d}_1, ..., \hat{d}_k\}$. In addition, we conduct keyword extraction over $q$ and $a$ to extract keywords from them. We then union the two sets of keywords (denoted by $W$).

\begin{figure*}[t]
    \centering
    \includegraphics[width=\linewidth]{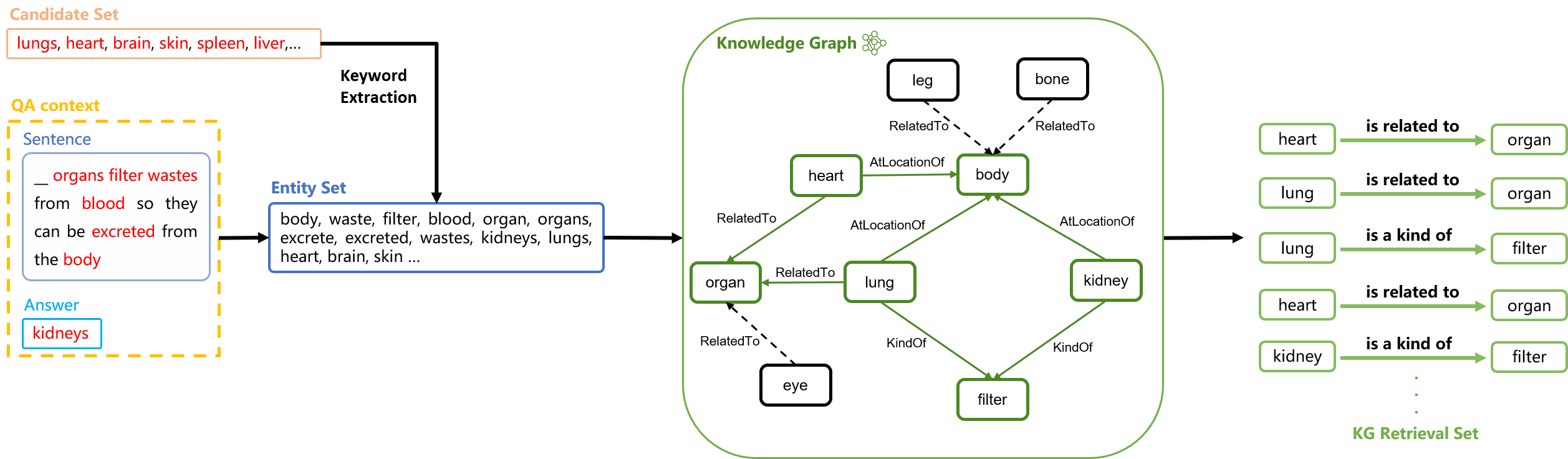}
    \caption{Retrieve Triplet from KG: we extract keyword from a given question, answer and candidate set as a entity set $W$ and retrieve relevant triplet set from KG with keyword entities}
    \label{fig:KG-Retrieval}
\end{figure*}

As illustrated in Figure \ref{fig:KG-Retrieval}, with $W$ and a given knowledge graph $G(V,E)$, the triplet set is given by
\[
K = \{(u, e_{u,v}, v) \mid u, v \in W \text{ and } e_{u,v} \in E\}
\]

\subsubsection{Triplet Ranker}\label{sec:ranker}

The input to triplet ranker is a triplet set $T$ from KG Retrieval stage and let $q||a$ be the concatenation of a question $q$ and an answer $a$.
We experiment with the following two alternatives for ranking triplets.
\begin{enumerate}
  \item \textbf{Unsupervised Sentence Embedding Distance}: The ranking process begins by encoding $q||a$ and each knowledge triplet $\tau_i$ into their respective vector representations. Next, we compute a relevancy score for each triplet by comparing its vector representation with that of $q||a$. The higher the relevancy score, the more likely the triplet is deemed relevant to $q||a$. Specifically,
\[
\text{{scores}} = [\text{{Cos-Similarity}}(\tau_i, q||a), \forall \tau_i \in K]
\]
We select top-$k$ triplets for the later knowledge augmented generation.





  \item \textbf{Supervised Triplet Classification}: As an alternative, we propose a binary classification formulation to estimate the relevancy of a knowledge triplet $\tau_i$ with respect to $q||a$. During the training phase, we label a triplet as relevant if it contains the answer or ground-truth distractors, and as irrelevant otherwise. Specifically, we concatenate $q||a$ and $\tau_i$, separated by a special token $[\mathtt{SEP}]$ as input and the relevancy as output. During the inference phase, we utilize the LM to re-sort the top-$k$ triplet set based on the confidence scores. 
  
  


\end{enumerate}

\subsubsection{KAG Training}
For a given training instance $(q, a, D, K)$, our goal is to train a generation model conditioned on $q$, $a$, and $K$ by minimizing the negative log-likelihood of the correct token $t_i$ of $D$ given the preceding tokens and the conditions.

\begin{itemize}
    \item $K$: the set of knowledge triplet $\{\tau_1, ..., \tau_k\}$ retrieved from a KG.
\end{itemize}

The idea is to retrieve a set of knowledge triplet $\{\tau_1, ..., \tau_k\}$ from a KG and concatenate the knowledge triplet with the original input text as an augmented text input for generation. Specifically, the loss function is
\[L(\theta) = -\sum_{i=1}^{|D|} t_{i} \log p(\hat{t}_{i} | \hat{t}_{<i}, q, a, \{\tau_i\}; \theta)\]

The observation behind the knowledge triplet augmentation strategy is to inject more information for generation through the knowledge graph in hope to boost the DG performance.






\section{Experiment}
\label{sec:experiment}

\subsection{Dataset}
We use SciQ \citep{welbl2017crowdsourcing} and MCQ dataset (the dataset releated by \citealp{ren2021knowledge}) for performance evaluation. For the details about the two benchmark datasets and our dataset setting. 

\paragraph{Sciq dataset} 
 The Sciq dataset is a multi-domain multiple-choice question dataset consisting of 13,000 questions in the fields of physics, chemistry, biology, and other natural sciences. These questions are open-ended and require reading and understanding relevant scientific knowledge, followed by reasoning and answering the questions. The questions are presented in a multiple-choice format, with each question accompanied by an answer and three incorrect options. According to the Sciq dataset configuration, the 13,000 data points are split into train/dev/test datasets with a ratio of 11,700/1,000/1,000.

Since the Sciq question sentences do not contain **blanks**, in the candidate generation method, we utilize generative LM, specifically ChatGPT, to generate the candidate set.

\paragraph{MCQ dataset}
MCQ dataset is a cross-domain cloze-style dataset, that includes the domains of science, vocabulary, common sense, and trivia.  Each data is composed of a sentence containing **blank** of cloze stem, answer, and distractors. According to the setting reported by \cite{Ren_Q.Zhu_2021}, MCQ contains 2880 questions and is randomly divided into train/dev/test with a ratio of 8:1:1. 


We obtain the MCQ dataset from GitHub link shared by \cite{Ren_Q.Zhu_2021}. However, we find there is a slight difference between the numbers in the shared dataset and reported in the paper. In the shared dataset, it only contains train and test data (with 2321/258). Thus, we use this data setting in our experiments. For dev data, we use 9:1 split from train as dev data.

\begin{table}
\centering
\resizebox{\linewidth}{!}{
\begin{tabular}{l|cccc|cccc}
\hline
\multirow{2}{*}{\textbf{Dataset}} & \multicolumn{4}{c}{\textbf{MCQ}} & \multicolumn{4}{|c}{\textbf{Sciq}}\\
\cline{2-9}
& \textbf{Train} & \textbf{Dev} & \textbf{Test} & \textbf{All} & \textbf{Train} & \textbf{Dev} & \textbf{Test} & \textbf{All}\\
\hline
{\# of Questions} & 2088 & 233 & 258 & 2580 & 11700 & 1000 & 1000 & 13700\\
\hline
\end{tabular}
}
\caption{\label{table:cloth-dataset}
The statistics of the training, development and test sets of MCQ, Sciq.
}
\end{table}

\subsection{Evaluation Metrics}
\paragraph{Automatic Metric}
Following the SOTA DG by \citep{juan-2023-PKLD}, we evaluate the DG quality by F1 score (F1@3), precision (P@1, P@3), and recall (R@1, R@3). P@k represents the ratio of correctly labeled top-k generated distractors, while R@k indicates the ratio of correctly predicted labels among the ground truth. F1@k is the harmonic mean of P@k and R@k. Notably, when the label size is 3, P@3 and R@3 will be the same, resulting in the same F1@3 score. Since both the Sciq test data and MCQ test data contain 3 distractors, we report the scores of P@1 and F1@3 in the experiments.

\paragraph{Human Evaluation Metric}
In addition to Automatic Metrics, we also consider human evaluation to assess the performance of the model on various aspects. During the evaluation process, we randomly select 14 questions and generate distractor sets using various models, which are then assessed by five human labelers. For the evaluation of the Distractor Generation models, we consider the following criteria for human evaluation:

\begin{enumerate}
\item \textbf{Relevance}: This metric assesses the relevance of the generated distractors to the question. This means that the distractors should be related to the theme, content, or context of the question.

\item \textbf{Distractiveness}: This metric evaluates the level of distraction that the generated distractors bring to the correct answer. The distractors should be able to divert the answer away from the correct answer, increasing the difficulty of the question.

\item \textbf{Utility}: This is an overall quality assessment of the generated distractors, considering aspects such as fluency, relevance, duplication, and distractiveness. This is a comprehensive metric that takes into account all facets of performance.
\end{enumerate}

We employ a five-point rating system, where each score interval represents different levels of relevance, ranging from "totally irrelevant" (1 point) to "completely relevant" (5 points). These evaluation metrics will help us assess the quality and performance of the distractors generated by the models in a comprehensive manner.

\subsection{Implementation Details}
See Appendix. 



\begin{table*}[t]
\centering
\resizebox{0.8\textwidth}{!}{
\begin{tabular}{c|l|ccccc}
\hline
\textbf{Dataset} & \textbf{Method} & \textbf{P@1} & \textbf{R@1} & \textbf{F1@3} & \textbf{MRR} & \textbf{NDCG@3}\\
\hline
\multirow{17}{*}{MCQ}
& \citep{ren2021knowledge} & 10.58 & - & 9.19 & 17.51 & -\\
& \citep{chiangCDGP} & 10.81 & 3.60 & 7.72 & 18.15 & 15.39\\
\cline{2-7}
& \citep{juan-2023-PKLD} (BART) & 14.28 & 4.76 & 11.45 & 21.49 & 23.70\\
& Text2text (RAP-BART) & 18.14 & 6.04 & 12.35 & 24.06 & 25.78\\
& \citep{juan-2023-PKLD} (BART w/ c.a.) & 19.69 & 6.56 & 13.12 & 25.03 & 26.26\\
& KAG (BART) & 20.07  & 6.69  & 14.41 & 27.02  & 29.37\\
& KAG (RAP-BART) & 16.60 & 5.75 & 12.22 & 22.65 & 24.50\\
& KAG (BART) (with only answer triplet) & 14.28 & 4.76 & 12.74 & 21.55 & 24.13 \\
& KAG (RAP-BART)(with only answer triplet) & 13.12 & 4.37 & 11.84 & 20.01 & 22.35 \\
\cline{2-7} 
& \citep{juan-2023-PKLD} (T5) & 18.53 & 6.17 & 11.45 & 23.61 & 25.08\\
& Text2text (RAP-T5) & \textbf{22.39} & \textbf{7.46} & 14.80 & \textbf{29.02} & 30.72\\
& \citep{juan-2023-PKLD} (T5 w/ c.a.) & 16.60 & 5.53 & 14.80 & 24.90 & 27.61\\
& KAG (T5) & 20.07& 6.69 & \textbf{16.47} & 28.57 & \textbf{30.99} \\
& KAG (RAP-T5) & 20.07 & 6.69 & 14.92 & 26.44 & 28.23 \\
& KAG (T5)(with only answer triplet) & 22.00 & 7.35 & 15.18 & 28.70 & 30.61\\
& KAG (RAP-T5)(with only answer triplet) & 21.62 & 7.20 & 14.41 & 27.22 & 28.60\\
\cline{2-7}
& {ChatGPT} & 18.91 & 6.30 & 13.38 & 25.86 & 27.97\\
\hline
\multirow{15}{*}{Sciq}
& \citep{juan-2023-PKLD} (BART) & 19.30 & 6.46 & 14.94 & 26.40 & 28.36\\
& Text2text (RAP-BART) & 20.10 & 6.73 & 16.32 & 27.56 & 29.68\\
& \citep{juan-2023-PKLD} (BART w/ c.a.) & 21.26 & 7.08 & 14.98 & 27.63 & 29.28\\
& KAG (BART) & 16.24 & 5.41 & 13.99 & 23.62 & 25.93\\
& KAG (RAP-BART) & 19.70 & 6.58 & 15.42 & 26.61 & 28.69\\
& KAG (BART) (with only answer triplet) & 17.15 & 5.57 & 14.87 & 25.00 & 27.27 \\
& KAG (RAP-BART)(with only answer triplet) & 18.35 & 6.15 & 15.52 & 25.81 & 28.22 \\
\cline{2-7}
& \citep{juan-2023-PKLD} Text2text (T5) & 24.27 & \textbf{8.09} & 15.92 & 30.32 & 31.99\\
& Text2text (RAP-T5)  & \textbf{24.30} & 8.09 & 16.06 & 29.98 & 31.49\\
& \citep{juan-2023-PKLD} (T5 w/ c.a.) & 22.06 & 7.35 & 12.83 & 25.54 & 26.47\\
& KAG (T5) & 22.50 & 7.51 & 15.38 & 29.10 & 31.04\\
& KAG (RAP-T5)  & 18.50 & 6.18 & 15.15 & 26.00 & 28.40\\
& KAG (T5)(with only answer triplet) & 23.70 & 7.91 & \textbf{16.50} & \textbf{30.41} & \textbf{32.39}\\
& KAG (RAP-T5)(with only answer triplet) & 23.00 & 7.68 & 16.46 & 29.51 & 31.41\\
\cline{2-7}
& {ChatGPT}  & 15.17 & 5.16 & 10.61 & 19.39 & 20.68\\
\hline
\end{tabular}
}
\caption{\label{table:cloth-distractor-generation}
DG Results on the Compared Datasets: In this set of experiment, we mainly compare our designs with the SOTA method \cite{juan-2023-PKLD}. c.a. represents candidate augmentation strategy introduced by \cite{juan-2023-PKLD}.  
}
\end{table*}




\begin{table}
\centering
\resizebox{\linewidth}{!}{
\begin{tabular}{c|l|c}
\hline 
Dataset                     &Method                                         &\# of masked Sentences/Passages  \\ \hline
\multirow{3}{*}{MCQ-train}  &RAP-S                                 &17,225  \\
                            &RAP-P                                 &17,362  \\  
                            &RAP-P    w/ GTD      &69,356   \\\hline
\multirow{3}{*}{MCQ-all}    &RAP-S                                  &10,894   \\
                            &RAP-P                                     &10,995   \\
                            &RAP-P    w/ GTD       &44,294   \\\hline
\multirow{3}{*}{Sciq-train} &RAP-S                                  &85,463   \\
                            &RAP-P                                     &86,269   \\
                            &RAP-P    w/ GTD       &307,915   \\  \hline
\multirow{3}{*}{Sciq-all}   &RAP-S                                  &91,798   \\
                            &RAP-P                                     &92,731   \\
                            &RAP-P     w/ GTD       &327,546   \\ \hline

\end{tabular}
}
\caption{\label{table:RAP trainin data}
The statistics of the RAP training. Note that our RAP has the following variations. First, in RAP, we can choose to retrieve a single sentence or a passage, giving rise to different RAP strategies. Thus, we denote RAP-S as retrieving sentence for pseudo question generation and RAP-P for retrieving passage. Second, we have the option to use the answer or the distractor as the starting point for retrieving the sentence/passage. We denote the method using distractor as starting point as RAP-S w/GTD or RAP-S/RAP-P (as using answer as starting point)
}
\end{table}

\begin{table*}[t]
\centering
\resizebox{0.9\textwidth}{!}{
\begin{tabular}{c|ll|ccccc}
\hline
\begin{tabular}{c}
     \textbf{DataSet for}\\ 
     \textbf{Traing and} \\
     \textbf{Testing DG model }
\end{tabular} & \multicolumn{1}{c}{\textbf{Method}} & \textbf{Pretrain Dataset} & \textbf{P@1} & \textbf{R@1} & \textbf{F1@3} & \textbf{MRR} & \textbf{NDCG@3} \\ \hline
\multirow{14}{*}{MCQ} & BART off-shelf & None & 14.28 & 4.76 & 11.45 & 21.49 & 23.70 \\ 
 & RAP-S$_{\text{BART}}$ & MCQ-Train & 16.21 & 5.4 & 11.96 & 22.2 & 23.99 \\
 & RAP-P$_{\text{BART}}$ & MCQ-Train & 18.14 & 6.04 & 12.35 & 24.06 & 25.78 \\
 & RAP-P$_{\text{BART}}$ /w GTD & MCQ-Train & 15.05 & 5.01 & 11.58 & 20.65 & 22.33 \\ \cline{2-8} 
 & RAP-S$_{\text{BART}}$ & Sciq-all & 15.05 & 5.01 & 13.25 & 22.77 & 25.44 \\
 & RAP-P$_{\text{BART}}$ & Sciq-all & 20.84 & 6.94 & 15.57 & 28.57 & 30.85 \\
 & RAP-P$_{\text{BART}}$ /w GTD & Sciq-all & 22.77 & 7.59 & 17.88 & 31.33 & 33.97 \\ \cline{2-8} 
 & T5 off-shelf & None & 18.53 & 6.17 & 11.45 & 23.61 & 25.08 \\  
 & RAP-S$_{\text{T5}}$ & MCQ-Train & 23.93 & 7.97 & 14.15 & 29.27 & 30.67 \\
 & RAP-P$_{\text{T5}}$ & MCQ-Train & 22.39 & 7.46 & 14.8 & 29.02 & 30.72 \\
 & RAP-P$_{\text{T5}}$ /w GTD & MCQ-Train & 21.62 & 7.2 & 13.77 & 27.15 & 28.70 \\ \cline{2-8} 
 & RAP-S$_{\text{T5}}$ & Sciq-all & 25.86 & 8.62 & 15.57 & 31.46 & 33.12 \\
 & RAP-P$_{\text{T5}}$ & Sciq-all & \textbf{31.66} & \textbf{10.55} & \textbf{18.91} & \textbf{37.7} & \textbf{39.43} \\
 & RAP-P$_{\text{T5}}$ /w GTD & Sciq-all & 27.79 & 9.26 & 18.66 & 34.42 & 36.31 \\ \hline
\multirow{14}{*}{Sciq} & BART off-shelf & None & 19.3 & 6.46 & 14.94 & 26.40 & 28.36 \\
 & RAP-S$_{\text{BART}}$ & Sciq-train & 19.3 & 6.45 & 14.66 & 25.61 & 27.45 \\
 & RAP-P$_{\text{BART}}$ & Sciq-train & 20.1 & 6.73 & 16.32 & 27.56 & 29.68 \\
 & RAP-P$_{\text{BART}}$ /w GTD & Sciq-train & 22.4 & 7.48 & 16.39 & 28.9 & 30.81 \\ \cline{2-8} 
 & RAP-S$_{\text{BART}}$ & MCQ-all & 20.4 & 6.83 & 15.65 & 27.19 & 29.09 \\
 & RAP-P$_{\text{BART}}$  & MCQ-all & 20.2 & 6.75 & 15.58 & 27.41 & 29.5 \\
 & RAP-P$_{\text{BART}}$ /w GTD & MCQ-all & 20.7 & 6.91 & 16.06 & 28.16 & 30.32 \\ \cline{2-8} 
 & T5 off-shelf & None & 24.27 & 8.09 & 15.92 & \textbf{30.32} & \textbf{31.99} \\ 
 & RAP-S$_{\text{T5}}$  & Sciq-train & 23.6 & 7.86 & 15.62 & 29.13 & 30.69 \\
 & RAP-P$_{\text{T5}}$  & Sciq-train & 24.3 & 8.09 & 16.06 & 29.98 & 31.49 \\
 & RAP-P$_{\text{T5}}$  /w GTD & Sciq-train & \textbf{25.0} & \textbf{8.34} & \textbf{16.46} & 30.18 & 31.54 \\ \cline{2-8} 
 & RAP-S$_{\text{T5}}$ & MCQ-all & 24.1 & 8.04 & 15.26 & 29.63 & 31.12 \\
 & RAP-P$_{\text{T5}}$ & MCQ-all & 23.1 & 7.71 & 16.31 & 29.56 & 31.49 \\
 & RAP-P$_{\text{T5}}$  /w GTD & MCQ-all & 23.7 & 7.89 & 16.36 & 29.73 & 31.25 \\ \hline
\end{tabular}
}
\caption{\label{table:RAP}
RAP Performance Overview with In-Domain and Cross-Domain Study.
}
\end{table*}


\subsection{Evaluation Results}

Table~\ref{table:cloth-distractor-generation} presents the results of the compared methods on the two benchmarking datasets. We have the following notes for the results. 

\noindent\textbf{\textit{RAP indeed improve performance}}
By comparing the performance difference between Text2Text (T5) and Text2Text (RAP-T5), we can see that the introduction of RAP elevates the performance of MCQ experiments from 11.45 to 14.80 (F1@3). In the case of Sciq, it increases from 15.92 to 16.06.

\noindent\textbf{\textit{RAP more effective in low resource settings}}
Furthermore, continuing the above observations, we found that the improvement brought by RAP in the Sciq experiment is not as significant as in the MCQ experiment. We attribute this result to the much smaller size of the MCQ dataset compared to Sciq. When the training data is abundant, the gains from pretraining naturally become limited. Therefore, in the MCQ setting with only 2088 training examples, the introduction of RAP has a noticeable impact.

\noindent\textbf{\textit{KG also boosts the performance}} By comparing T5 candidate augmentation with KAG (T5), we observe that the introduction of KG indeed brings a significant improvement. In Sciq, we observe an increase in F1@3 from 12.83 to 15.38. In MCQ, it increases from 14.80 to 16.47.

\noindent\textbf{\textit{Our methods outperform ChatGPT}} We also compare the performance of ChatGPT (in a zero-shot manner) on both datasets. From the experimental results, it is evident that our best-performing models consistently outperform ChatGPT. In the comparison with Sciq, KAG (RAP-T5) achieves an F1@3 score of 15+ compared to ChatGPT's 10.61. An interesting observation is that while ChatGPT still lags behind the KAG+RAP-based approach in the MCQ comparison, the performance gap is smaller. We attribute this to the fact that some questions in Sciq require common-sense reasoning in addition to knowledge-based answering. Since ChatGPT generates responses in a zero-shot learning manner during testing, it does not fully capture the characteristics of the Sciq dataset. In the case of the MCQ dataset, due to the limited training data, the differences between various methods and ChatGPT are not as substantial as in the Sciq comparison.

\noindent\textbf{\textit{Combining KAG with RAP did not yield additive effects}} We also observed that combining KAG with RAP did not yield additive effects. Taking the comparison of the T5 model on the Sciq dataset as an example, we found that the performance did not improve when using both KAG and RAP together (KAG(RAP-T5): 15.15) compared to using RAP alone (Text2Text(RAP-T5): 16.06) or KAG alone (KAG(T5): 15.38). Similar observations were made in other settings. We speculate that the additional noise introduced by KAG during the incorporation of knowledge triplets may be the reason behind this. Although we designed a reranker to select triplets, the results were not consistently stable. To validate this hypothesis, we attempted a variation of KAG (RAP-T5) (denote by with-only-answer triplet) where, instead of using a reranker, we selected triplets containing the answer to input into the Text2Text generator. We observed significant improvement with this approach in the Sciq dataset, but no improvement in the MCQ dataset. This suggests that there is considerable room for improvement in the design of the triplet ranker. Selecting relevant triplets for generation is an area that warrants further exploration in future research.

\begin{table}
\centering
 \resizebox{\linewidth}{!}{
\begin{tabular}{l|ccc}
\hline
\textbf{Method} & \textbf{Relevance} & \textbf{Distractiveness} & \textbf{Utility} \\
\hline
\citep{juan-2023-PKLD} & 3.81 & 3.35 &  3.55\\
KAG(T5) &  4.05 & 3.25 & 3.77\\
Text2text(RAP-T5) &  \textbf{4.45} & \textbf{3.87} & \textbf{4.07}\\
{ChatGPT} & 3.14 & 2.67 & 3 \\
\cline{1-4}
{ground truth} & 3.52 & 3.02 & 3.4\\
\hline
\end{tabular}
 }
\caption{\label{table:human-evaluation}
The human evaluation for MCQ dataset. 
}
\end{table}

\paragraph{Human Evaluation Results}

Table \ref{table:human-evaluation} presents the results of various methods in generating distractor sets for the MCQ cloze tasks. Upon analysis, we observe that in terms of Relevance, Distractiveness, and Utility, both RAP and KAG scored relatively high and exceeded the previous soft method, T5 Candidate Augmentation.

Table \ref{table:human-evaluation} presents the results of various methods in generating distractor sets for the MCQ cloze tasks. Upon analysis, we observe that in terms of Relevance, Distractiveness, and Utility, both RAP and KAG scored high and surpassed or comparable to the prior SOTA method.

These results suggest that we have successfully guided our model to leverage additional knowledge to generate higher quality distractors. The reason why RAP outperforms KAG might be that the KG (ConceptNet) cannot encompass knowledge from all natural domains, whereas RAP, having access to an extensive external corpus (Wikipedia), includes a vast array of facts across various domains. This vastness could potentially explain why KAG scored slightly lower than RAP.

Furthermore, we noted that the ground truth quality for MCQ questions was not consistently high. The lowest performance was found in Distractiveness. This could likely be attributed to the fact that some of MCQ questions are generally designed for elementary and middle school grade levels, which could impact the complexity and misdirection quality of the distractors.



\begin{table}[t]
\centering
\begin{tabular}{l|l}
\hline
\textbf{Method} & \textbf{F1@3}\\
\hline
T5 KAG & \textbf{16.47}\\
\hline
w/o c.a & 13.64\\
w/o c.a (Masked LM) & 15.18\\
w/o c.a. (Generative LM) & 13.89\\
w/o reranker & 14.15\\
\hline
\end{tabular}
\caption{\label{table:ablation-study}\textbf{Ablation-study} of our KAG model components using the MCQ Test Dataset} \vspace{-2mm}
\end{table}

\subsection{Ablation Study on KAG}
We conducted ablation experiments on the MCQ dataset to assess the impact of different components in our system. The results are summarized in Table \ref{table:ablation-study}, providing insights into the importance and contributions of each component.

\paragraph{Candidate Generation}
Generating candidate sets using Masked LM or Generative LM is a key component in our system, enhancing KG retrieval. Removing these sets and relying solely on sentence and answer-based triplet retrieval significantly reduced the number of retrievable triplets, potentially introducing irrelevant noise and resulting in a performance drop (F1@3: 16.47 to 13.64). Using Masked LM or Generative LM alone also affected F1@3. This emphasizes the value of incorporating candidate sets, as they provide crucial information for identifying important triplets in the knowledge graph.

\paragraph{Reranker}
The Reranker, our second key component, sorts triplets to select those more relevant to distractors/answers. Without sorting by the Reranker, randomly selecting triplets for triplet augmentation led to a decrease in F1@3 (16.47 to 14.15). This highlights the Reranker's importance in identifying important triplets and improving downstream task performance.


\subsection{Study on Cross-Domain for RAP}\label{sec:cross}
In this subsection, we report our studies on cross-domain and in-domain for RAP. The experiment results are summarized in Table \ref{table:RAP}.

We have the following observations: Firstly, in terms of cross-domain effects (using the MCQ dataset for pretraining and fine-tuning the DG model with the Sciq dataset, or vice versa), we found that using Sciq-all for pretraining the language model resulted in a significant improvement in the downstream MCQ dataset for distractor generation. Specifically, using the T5 model, the RAP-based approach increased the F1@3 score from 11.45 to 18.91 (almost a 1.5-fold improvement). This notable improvement was expected, considering that Sciq is a larger dataset compared to MCQ. However, when we reversed the process and used MCQ for pretraining, followed by testing with the Sciq dataset, the improvement in performance was not as significant. Taking the BART model as an example, the direct use of BART off-shelf achieved a score of 14.94, while RAP-S$_{\text{BART}}$) only improved it to 16.06. Although there was some improvement, it was not as substantial as the effects observed when using Sciq pretraining.

Regarding in-domain effects (using the same dataset for both pretraining and fine-tuning), we found consistent improvements in the generated results. For example, using the T5 model on the Sciq test dataset, the T5 off-shell score was 15.92, while RAP-P$_{\text{T5}}$) improved it to 16.46. We also observed similar improvements in the MCQ dataset.

In conclusion, our proposed RAP approach has shown improved generation performance from both data augmentation and task-specific pretraining perspectives. The experiments demonstrate that incorporating RAP into the training process leads to significant enhancements in distractor generation.


\section{Conclusion}
\label{sec:conclusion}

In this paper, we introduce the utilization of task-specific pretraining and knowledge base. Our experimental results highlight a significant performance improvement achieved through the integration. our work represents a novel contribution to the field of distractor generation for MCQs, showcasing the potential of combining retrieval augmented pretraining and KGs to achieve superior results.


\section{Limitations}
We report the following limitations for the KAG method:

\begin{itemize}
    \item Combining KAG with RAP did not yield additive effects. We believe that the key factor lies in that the  knowledge triplets may introduces additional noise that interferes with the Text2Text model's generation process. Thus, there is significant room for improvement in the current design of the triplet ranker. Selecting relevant triplets for generation is an aspect that requires further strengthening in this research.
    \item The current evaluation and training heavily rely on token scores, which can only reflect the similarity to the ground truth but cannot fully represent the quality of the generated output. The research conclusions thus far have also been established solely based on token scores.
    \item During the pretraining stage of RAP, when retrieving sentences/passages, if the answer is a rare or specialized term, the external corpus may fail to find matching sentences/passages. Therefore, when applying the RAP framework to other knowledge domains, having an abundant knowledge corpus becomes crucial.
   
\end{itemize}


\section*{Acknowledgement}
This work is supported by NSTC 112-2634-F-005 -002-project Smart Sustainable New Agriculture Research Center (SMARTer), NSTC Taiwan Project under grant 112-2221-E-005 -075 -MY3, and Delta Research Center, Delta Electronics, Inc.


\bibliography{anthology}
\bibliographystyle{acl_natbib}

\appendix
\noindent\textbf{\large Appendix}

\section{Implementation Details}
Our models are implemented based on models from Hugging Face \citep{wolf2019huggingface}. 
For generating candidate set, we use an language model trained for distractor
generation from \citep{chiangCDGP} and gpt3.5 turbo through ChatGPT API with zero-shot prompting. For triplet reranker, we use Sentence-BERT \citep{reimers2019sentence} to estimate the initial sentence embedding distance. For triplet classification, we use Sentence-BERT and BERT \citep{devlin2018bert} as the default PLM. For the triplet augmented generator, we experiment with BART \citep{lewis2019bart} and T5 \citep{raffel2020exploring} as base generation models.

During training, we use AdamW as the optimizer and an initial learning rate of 2e-5 for BERT, BART and 1e-4 for T5 models. All experiments are conducted using two NVIDIA TITAN RTX GPUs.


\paragraph{BART-based generator} 
With MCQ data, the maximum number of epochs is set to 40 with a batch size of on two NVIDIA TITAN RTX GPUs for the distraction generation with triplet augmentation with a batch size of 32 and a maximum number of triplets is set to 50. With Sciq data, the maximum number of epochs is set to 50 with a batch size of 32 on two NVIDIA TITAN RTX GPUs for the distraction generation with triplet augmentation with a batch size of 32 and a maximum number of triplets is set to 50. The average running time for BART-based generators is 30 minutes (1.5 hours) on MCQ (Sciq).


\paragraph{T5-based generator} 
With MCQ data, the maximum number of epochs is set to 40 with a batch size of on two NVIDIA TITAN RTX GPUs for the distraction generation with triplet augmentation with a batch size of 32 and a maximum number of triplets is set to 50. With Sciq data, the maximum number of epochs is set to 50 with a batch size of 32 on two NVIDIA TITAN RTX GPUs for the distraction generation with triplet augmentation with a batch size of 32 and a maximum number of triplets is set to 50. The average running time for BART-based generators is 40 minutes (3 hours) on MCQ (Sciq).

\paragraph{Knowledge Graph} We utilize ConceptNet \citep{speer2017conceptnet} as our KG, which is a graph that encompasses a wide range of knowledge domains. The KG consists of a total of 800k nodes and 2M edges. For each question's KG retrieval set, we preprocess the retrieval subgraph using the approach described in \citep{feng2020scalable}. Specifically, we process the textual corpus of each question, including the question text, answer options, and entities in the candidate set.

\end{document}